# Rain structure transfer using an exemplar rain image for synthetic rain image generation

Chang-Hwan Son and Xiao-Ping Zhang

{changhwan76.son@gmail.com; xzhang@ryerson.ca}

Ryerson University, Toronto, Canada

**Abstract:** This letter proposes a simple method of transferring rain structures of a given exemplar image (rain image) into a target image (non-rain image). Given the exemplar rain image and its corresponding masked rain image, rain patches including real-life rain structures are extracted randomly, and then residual rain patches are obtained by subtracting those rain patches from their mean patches. Next, residual rain patches are selected randomly, and then added to the given target image along a raster scanning direction. To decrease boundary artifacts around the added patches on the target image, minimum error boundary cuts are found using dynamic programming, and then blending is conducted between overlapping patches. Our experiment shows that the proposed method can generate realistic rain images that have similar rain structures in the exemplar images. Moreover, it is expected that the proposed method can be used for rain removal. More specifically, non-rain images and synthetic rain images generated via the proposed method can be used to learn classifiers (e.g., deep neural network) in a supervised manner.

**1. Introduction:** Rain forms structures on captured images. This means that rain structures can prevent computer vision algorithms (e.g., car/sign detection, visual saliency, scene parsing, etc.) from working effectively [1]. Most computer vision algorithms depend on feature descriptors such as scale invariant feature transform (SIFT) [2] and histogram of oriented gradients (HOG) [3]. These descriptors are designed based on the gradient's magnitude and orientation, and thus rain structures can have negative effects on the feature extractor. For this reason, rain removal is a necessary tool [4]. However, to learn classifiers (e.g., deep neural network) in a supervised manner, it is necessary to collect rain and clean patch pairs. This requires synthetic rain image generation. There are various types of methods [5-9] that transfer colors, textures, or styles of the exemplar images into the target images. Different from the conventional methods, in this letter, rain structure transfer method is presented.

**2. Proposed Algorithm:** The detail procedure of generating synthetic rain images is as follows:

- Prepare an exemplar rain image and a target non-rain image
- Define rain regions in the exemplar image by using a brush tool in Abode Photoshop.
- Extract rain patches from the classified rain regions in the exemplar image
- Repeat the following steps:
    o Select one rain patch from the extracted rain patches randomly
    o Obtain the residual rain patch from the selected rain patch with the following equation: $\mathbf{P}^o = \mathbf{P}^r - \text{mean}(\mathbf{P}^r)$ where $\mathbf{P}^r$ is the selected rain patch and $\text{mean}(\cdot)$ is a function that produces a mean vector from an input vector
    o Add the residual rain patch ($\mathbf{P}^o$) to the non-rain patch ($\mathbf{P}^t$) at the $(i,j)$ block position in the target image, thereby producing synthetic rain patch ($\mathbf{P}^s = \mathbf{P}^o + \mathbf{P}^t$)
    o Find the minimum error boundary cut [8] using dynamic programming for the rain patch ($\mathbf{P}^r$), and then blend overlapping areas between the synthetic rain patches by using the minimum error boundary cut
    o Move to the next block position in the target image
- Until current block position $(i,j)$ reaches the last block position

**3. Experimental Results:** To find the minimum error boundary cut [8], we used the source code provided by authors. The proposed method was coded with Matlab (R2016a), and then tested on Window 7. As shown in Figs. 1-4, the proposed method can generate realistic rain images by transferring real-life rain structures of the exemplar images into the target images.

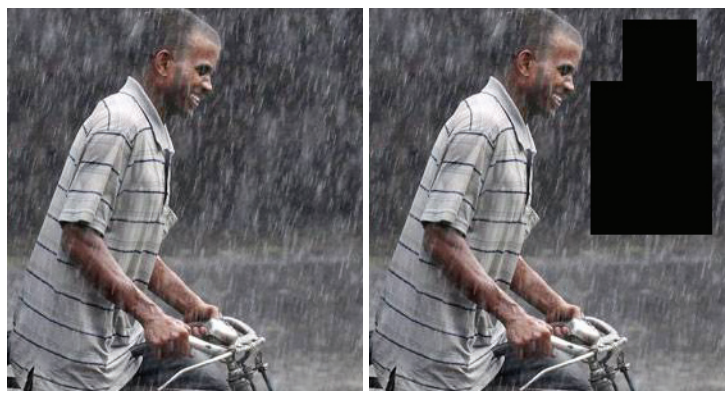

(a)                    (b)

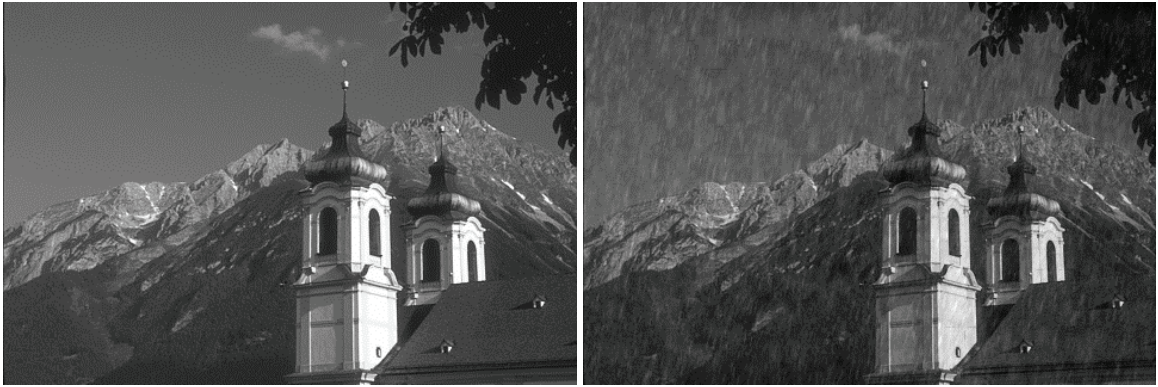

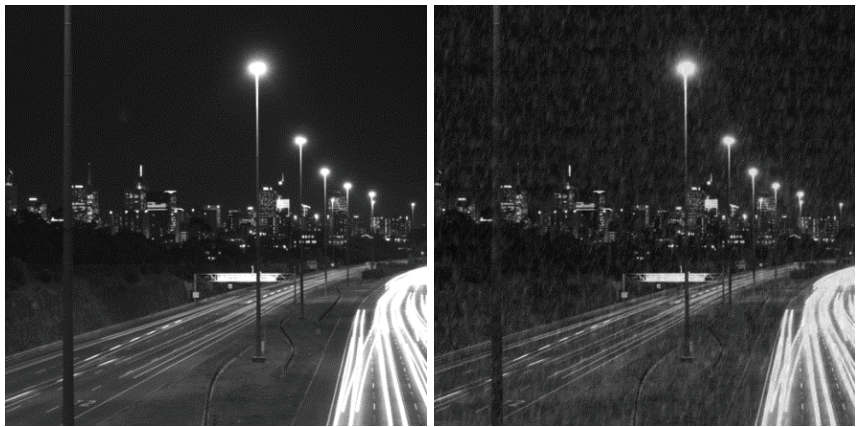

Fig. 1. Resulting images: (a) an exemplar 'rain' image, (b) masked 'rain' image, (c) target 'non-rain' image, (d) synthetic rain image generated via the proposed rain structure transfer, (e) target 'non-rain' image, and (f) synthetic rain image generated via the proposed rain structure transfer.

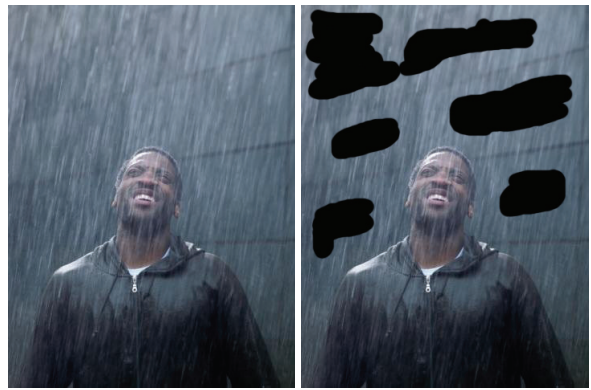

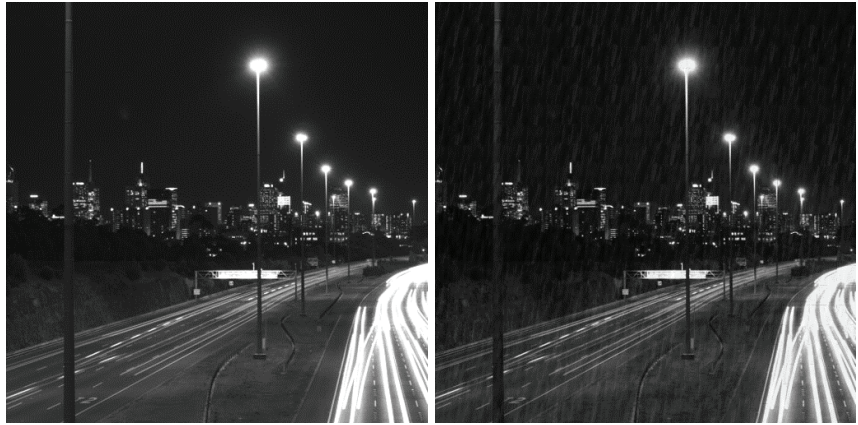

(c)　　　　　　　　　　　　　(d)

Fig. 2. Resulting images: (a) an exemplar 'rain' image, (b) scribbled 'rain' image, (c) target 'non-rain' image, and (d) synthetic rain image generated with the proposed rain structure transfer.

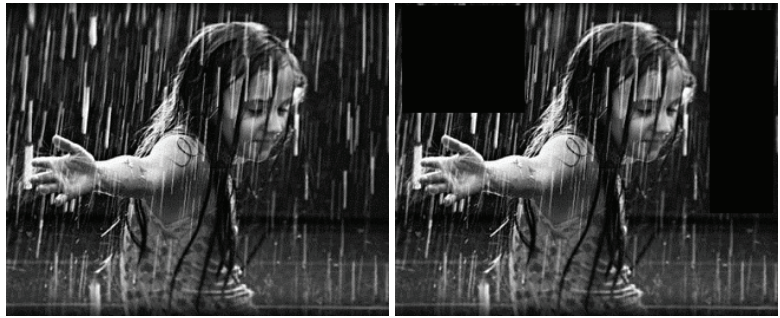

(a)　　　　　　　　　　　　　(b)

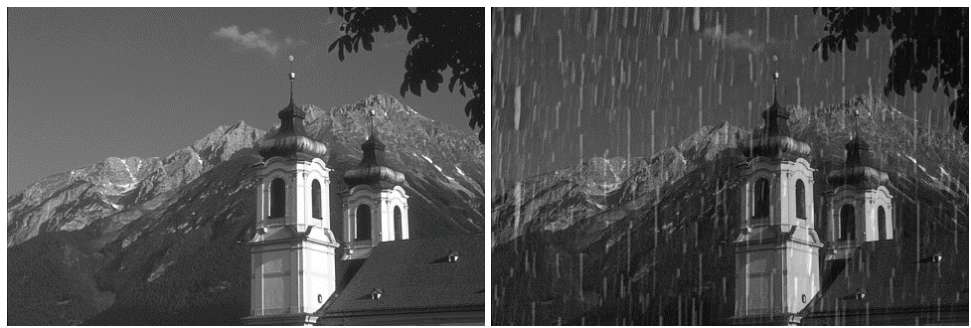

(c)　　　　　　　　　　　　　(d)

Fig. 3. Resulting images: (a) an exemplar 'rain' image, (b) masked 'rain' image, (c) target 'non-rain' image, and (d) synthetic rain image generated via the proposed rain structure transfer.

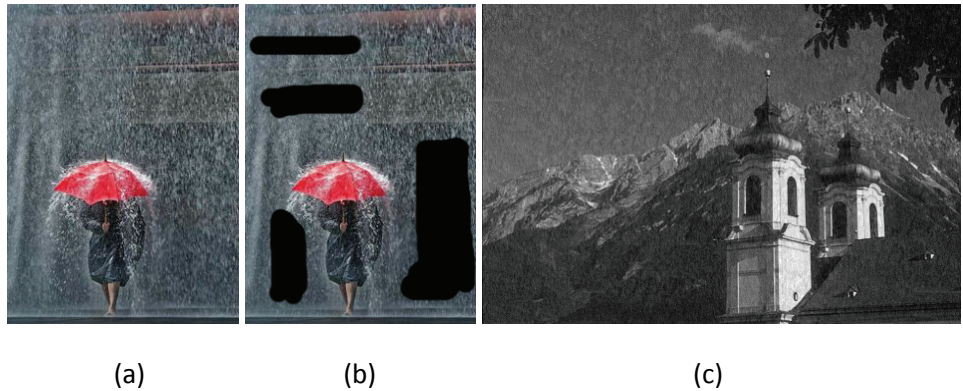

(a) (b) (c)

Fig. 4. Resulting images: (a) an exemplar 'rain' image, (b) scribbled 'rain' image, and (c) synthetic rain image generated via the proposed rain structure transfer.

**4. Implementation about rain removal:** Most regression models using learned parameters (e.g., deep neural network) require patch pairs. Therefore, in the proposed method, it is unnecessary to blend overlapping areas between synthetic rain patches for rain removal. In other words, to learn classifiers, only extracted non-rain patches and the corresponding synthetic rain patches are needed.

**5. Conclusion:** In this letter, rain structure transfer method is proposed. To achieve this, exemplar rain images are used to transfer real-life rain structures into the given target images. Our experiment shows that the proposed method can produce realistic rain images that have similar rain structures in the given exemplar rain images.